\documentclass{article}

\usepackage{microtype}
\usepackage{graphicx}
\usepackage{subfigure}
\usepackage{booktabs} % for professional tables

\usepackage{enumitem}
\usepackage{amsfonts}
\usepackage{amssymb}
\usepackage{amstext}

\usepackage{amsmath}
\usepackage{xspace}

\newcommand{\datainstance}{\boldsymbol{x}}
\newcommand{\labelinstance}{y}

\newcommand{\dataset}{\mathcal{D}}

\newcommand{\numdatapoints}{N}
\newcommand{\model}{f}
\newcommand{\obj}{G}

\newcommand{\counterfactual}{\textrm{cf}}
\newcommand{\alg}{\mathcal{A}}
\newcommand{\cost}{d}

\newcommand{\pro}{\textrm{pr}}
\newcommand{\unpro}{\textrm{np}}

\newcommand{\posout}{\textrm{pos}}
\newcommand{\negout}{\textrm{neg}}

\newcommand{\key}{\boldsymbol{\delta}}

\newcommand{\mparams}{\boldsymbol{\theta}}

\newcommand{\scaffolding}{scaffolding}

\usepackage{hyperref}

\usepackage[accepted]{icml2021}

\icmltitlerunning{Feature Attributions and Counterfactual Explanations can be Manipulated}

\begin{document}

\twocolumn[
\icmltitle{Feature Attributions and Counterfactual Explanations Can Be Manipulated}



\icmlsetsymbol{equal}{*}

\begin{icmlauthorlist}
\icmlauthor{Dylan Slack}{ir}
\icmlauthor{Sophie Hilgard}{har}
\icmlauthor{Sameer Singh$_{\dagger}^1$}{}
\icmlauthor{Himabindu Lakkaraju$_{\dagger}^2$}{}
\end{icmlauthorlist}

\icmlaffiliation{ir}{University of California, Irvine}
\icmlaffiliation{har}{Harvard University. $\dagger$ equal senior author contribution. \textit{ICML 2021 workshop on Theoretic Foundation, Criticism, and Application Trend of Explainable AI}}
% \icmlaffiliation{ed}{School of Computation, University of Edenborrow, Edenborrow, United Kingdom}

% \icmlcorrespondingauthor{Cieua Vvvvv}{c.vvvvv@googol.com}
% \icmlcorrespondingauthor{Eee Pppp}{ep@eden.co.uk}

\icmlkeywords{Machine Learning, ICML}

\vskip 0.3in
]



\printAffiliationsAndNotice{}  % leave blank if no need to mention equal contribution

\begin{abstract}
As machine learning models are increasingly used in critical decision-making settings (e.g., healthcare, finance), there has been a growing emphasis on developing methods to explain model predictions.
Such \textit{explanations} are used to understand and establish trust in models and are vital components in machine learning pipelines.
Though explanations are a critical piece in these systems, there is little understanding about how they are vulnerable to manipulation by adversaries.
In this paper, we discuss how two broad classes of explanations are vulnerable to manipulation.
We demonstrate how adversaries can design biased models that manipulate model agnostic feature attribution methods (e.g., LIME \& SHAP) and counterfactual explanations that hill-climb during the counterfactual search (e.g., Wachter's Algorithm \& DiCE) into \textit{concealing} the model's biases.
These vulnerabilities allow an adversary to deploy a biased model, yet explanations will not reveal this bias, thereby deceiving stakeholders into trusting the model.  
We evaluate the manipulations on real world data sets, including COMPAS and Communities \& Crime, and find explanations can be manipulated in practice.
\end{abstract}

\section{Introduction}
Recently, there has been considerable interest in leveraging machine learning (ML) models in critical applications. Consequently, it becomes crucial to ensure that the predictions made by these models are understood and trusted by domain experts (e.g., doctors) and other relevant stakeholders. 
However, both the proprietary nature and complexity of ML models make it challenging for domain experts to understand them. This behavior motivates the need for tools that can explain models in a faithful and interpretable manner. To this end, several classes of post hoc explanation methods---e.g.,
local~\cite{lime,shap-lunberg,smilkov2017smoothgrad,sundararajan2017axiomatic}, global~\cite{lakkaraju16:interpretable, letham15:interpretable}, prototype/exemplar-based~\cite{NEURIPS2019_adf7ee2d}, and counterfactual explanation~\cite{wachter,proto20} methods---have been proposed.

Such post hoc explanations are used in critical settings to detect discriminatory biases in black box models and decide if such models are trustworthy~\cite{umangspaper}. Thus, it is crucial to ensure that post hoc explanations are reliable, capture the overall behavior and other critical properties (e.g., fairness) of the underlying models accurately. An initial step towards this goal is to rigorously analyze the behavior of state-of-the-art post hoc explanation methods and identify their vulnerabilities. Recent research has focused on identifying such shortcomings. For example,~\citet{adebayo2018sanity} demonstrated that the explanations output by gradient-based methods are not faithful to the underlying model. Furthermore,~\citet{AlvarezMelis2018OnTR} demonstrated that the explanations output by popular techniques such as LIME and SHAP are unstable, i.e., infinitesimally small changes to instances can cause their post hoc explanations to change drastically.

While prior work has demonstrated some of the vulnerabilities of existing post hoc explanation methods, there is little understanding as to whether adversaries can manipulate explanations and mislead domain experts into trusting clearly untrustworthy models. Understanding this question is crucial because it both underscores the risks of reliance on post hoc explanations and encourages practitioners to exert caution in what they infer from these methods. In this work, we demonstrate how feature attribution methods and counterfactual explanations are vulnerable to manipulation.
Specifically, we show that it is possible for adversaries to design models where feature attribution methods return \textit{any} desired explanation, allowing clearly biased models to look unbiased in explanations.
We also demonstrate that it is possible to design models where counterfactual explanations appear unbiased but return much easier to obtain counterfactuals under a slight perturbation, allowing an adversary to selectively perturb instances to generate low cost recourse, thereby introducing hidden biases.
We experiment on real world data sets using attribution methods, such as LIME \cite{lime}, and counterfactual explanation methods, including Wachter's algorithm \cite{wachter} and DiCE \cite{mothilal2020dice} to show the efficacy of the manipulations.

\section{Preliminaries}

In this section, we introduce notation and provide background on the explanation methods we consider. 
% We also describe the our general problem setup for manipulating post hoc explanations.

\subsection{Notation}

We use a dataset $\dataset$ %containing binary protected attributes and labels. 
containing $\numdatapoints$ data points, where each instance is a tuple of $\datainstance \in \mathbb{R}^d$ and label $\labelinstance \in \{0,1\}$, i.e.
% \sameer{looks like 1 attribute? is it in $x$? maybe a function of x?} 
% The training set takes on the form 
$\dataset = \{ (\datainstance_n, \labelinstance_n ) \}_{n=1}^{N}$ (similarly for the test set).  
For convenience, we refer to the set of all data points $\datainstance$ in dataset $\dataset$ as $\mathcal{X}$. 
% We will use the notation $\datainstance_i$ to denote indexing data points in $\mathcal{X}/\dataset$ and $\datainstance^q$ to denote indexing attribute $q$ in $\datainstance$.
% , and sensitive attribute $\protectedinstance \in \{0,1\}$.
% We also assume the dat
Further, we have a model that predicts the probability of the positive class using a datapoint $\model : x \rightarrow [0,1]$.  
% Further, we assume the model is paramaterized by $\mparams$ but omit the dependence and write $f$ for convenience. 
We assume the positive class is the desired outcome (e.g., receiving a loan) henceforth. Last, we also assume we have access to whether each instance in the dataset belongs to a protected group of interest or not, to be able to define fairness requirements for the model.
% The protected group refers to an historically disadvantaged group such as women or African-Americans.   
We use $\dataset_{\pro}$ to indicate the protected subset of the dataset $\dataset$, and use $\dataset_{\unpro}$ for the ``not-protected'' group.
Further, we denote protected group with the positive (i.e. \emph{more desired}) outcome as $\dataset_{\pro}^{\posout}$ and with negative (i.e. \emph{less desired}) outcome as $\dataset_{\pro}^{\negout}$ (and similarly for the non-protected group). 

\subsection{Model Agnostic Feature Attribution Explanations}

We will focus on the class of model agnostic feature attribution explanations that fit an \textit{interpretable} local linear model around a prediction in order to explain a complex black box.
LIME~\cite{ribeiro16:kdd} is one such popular \emph{model-agnostic}, \emph{feature attribution} approach. This method explains individual predictions by learning a linear model, $g$, locally around each prediction.
Specifically, LIME estimates feature attributions on individual instances, which capture the \emph{contribution} of each feature on the black box prediction.
% Below, we provide some details of these approaches, while also highlighting how they relate to each other.
LIME samples a set of \textit{perturbations} in the neighborhood of a point and fits a weighted regression, where weighting is determined by distance to the point.
% $\pi_{x}(x')$. The objective for LIME \& SHAP is given as,
% \[L(f, g, \pi_{x})  = \sum_{x' \in X'} [f(x') - g(x')]^2 \pi_{x}(x')\]
% where $X'$ is the set of inputs constituting the neighborhood of $x$.
% The main difference between LIME \& SHAP is how the weighting function is chosen. 
% LIME uses $\ell_2$ or cosine distance while SHAP uses a function determined by game theoretic principles that satisfies certain desirable properties. 

\subsection{Counterfactual Explanations}

Counterfactual explanations return a data point that is \emph{close} to $\datainstance$ but is predicted to be positive by the model $\model$.
% \sameer{clarify $f(x)<0.5$? some other $\tau$? we want it to be $f(A(x))>\tau$?} 
We denote the counterfactual returned by a particular algorithm $\alg$ for instance $\datainstance$ as $ \alg (\datainstance)$.
% We take the difference between the original data point $\datainstance$ and counterfactual $\alg (\datainstance)$ as the set of changes an individual would have to make to receive the desired outcome. 
% We refer to this set of changes as the \textit{recourse} afforded by the counterfactual explanation.  
We define the \textit{cost} of recourse as the \textit{effort} required to achieve the counterfacutal~\cite{philosophicalbasis}. 
% In this work, we define the cost of recourse as the distance between $\datainstance$ and $\alg (\datainstance ) $.   
% Because computing the real-world cost of recourse is challenging \cite{hiddenassumption}, we use an ad-hoc distance function, as is general practice.
In general, counterfactual explanation techniques optimize objectives of the form,
\begin{align}
        \obj (\datainstance, \datainstance_\counterfactual) & =  \lambda \cdot \left( \model(\datainstance_\counterfactual) - 1 \right)^2 + \cost(\datainstance, \datainstance_\counterfactual)
        % + \reg(\datainstance_\counterfactual) 
        \label{eq: G}
        \\
       \alg(\datainstance) & = \underset{{\datainstance_\counterfactual}}{\textrm{argmin}} \; \obj (\datainstance, \datainstance_\counterfactual)
        \label{eq: counterfactual general form}
\end{align}
where $\datainstance_\counterfactual$ denotes \textit{candidate} counterfactual at a particular point during optimization. %, i.e., the counterfactual
The first term $\lambda \cdot \left( \model(\datainstance_\counterfactual) - 1 \right)$ encourages the counterfactual to have the desired outcome probability by the model. 
The distance function $\cost(\datainstance, \datainstance_\counterfactual)$ enforces that the counterfactual is close to the original instance and easier to ``achieve'' (lower cost recourse) and varies by the counterfactual explanation, so as to achieve different desired behaviors.
$\lambda$ balances the two terms. 
% As one such notion of distance, \citet{wachter} use the Manhattan ($\ell_1$) distance weighted by the inverse median absolute deviation (MAD).
% \begin{align}
%     & \cost(\datainstance, \datainstance_\counterfactual)  = \sum_{q\in [d]} \frac{|\datainstance^q - \datainstance_\counterfactual^q|}{\textrm{MAD}_q} \\
%     \textrm{MAD}_q & = \textrm{median}_{i \in [N]} \left( | x^q_i - \textrm{median}_{j \in [N]} (x_j^q) |\right)  
% \end{align}
% This distance function generates sparse solutions and closely represents the absolute change someone would need to make to each feature, while correcting for different ranges across the features. 
% This distance function $\cost$ can be extended to capture other counterfactual algorithms, e.g., by adding a terms encouraging diversity as in DiCE~\cite{mothilal2020dice}.
We refer to the class of counterfactual explanations that optimize the counterfactual objective through gradient descent or black-box optimization as those that \textit{hill-climb} the counterfactual objective (e.g., Wachter's algorithm \cite{wachter} or DiCE \cite{mothilal2020dice}).
% fit this characterization because they optimize objective \ref{eq: counterfactual general form} through gradient descent. Techniques like MACE \cite{mace20} and FACE \cite{face} do not fit this criteria because they do not use hill-climbing techniques.

\section{Manipulating Explanations Overview}

\begin{figure}
    \centering
    \includegraphics[scale=0.45,clip,trim=0 0 0 40]{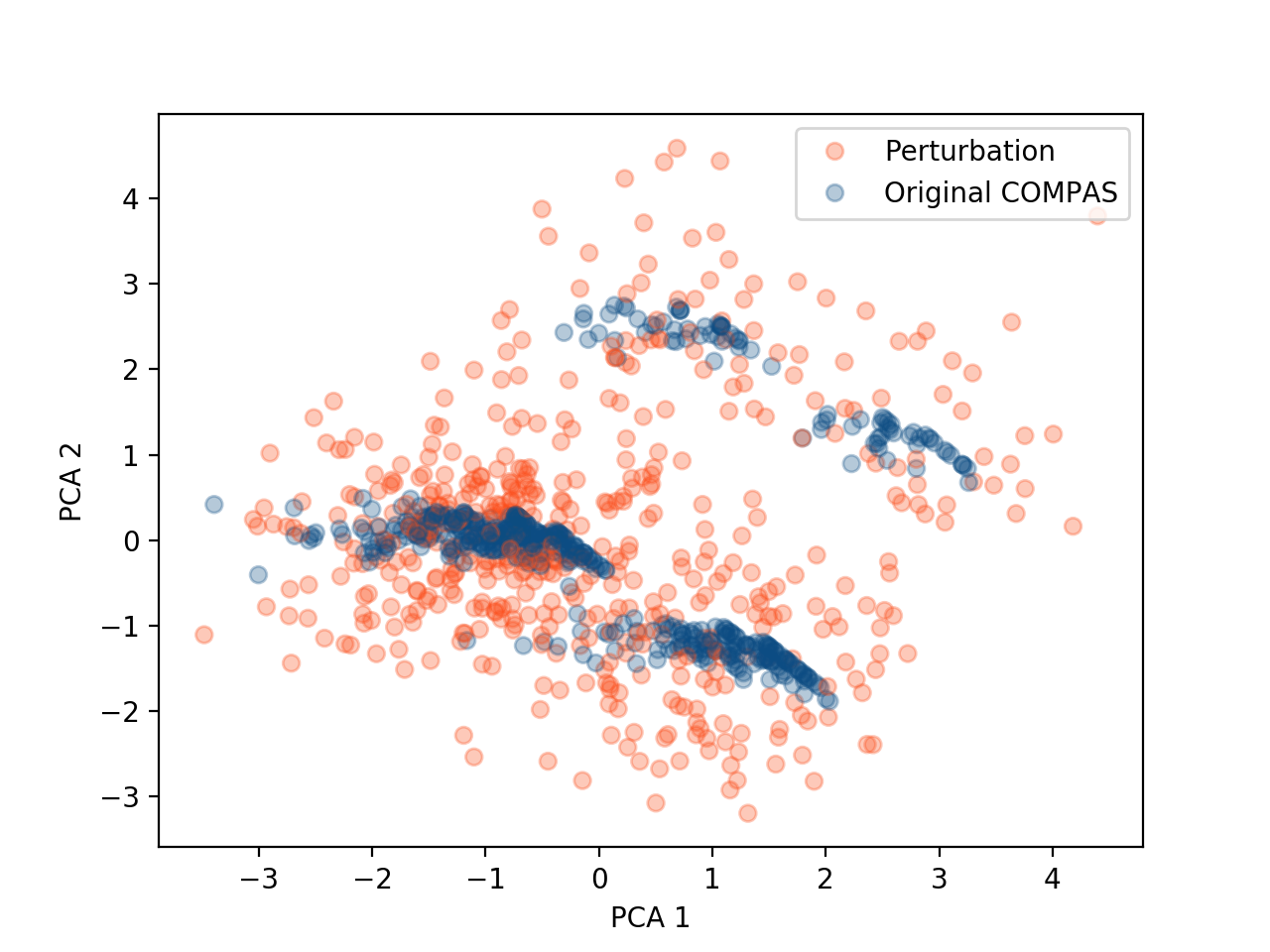}
    %\caption{PCA applied to the COMPAS dataset (blue) as well as its LIME style perturbations (red).  Even in this low-dimensional space, we can see that there is a significant difference between the perturbations and the true data distribution. In this paper, we exploit this difference to craft adversarial classifiers.}
    \caption{PCA applied to the COMPAS dataset (blue) as well as its LIME style perturbations (red).  Even in this low-dimensional space, we can see that data points generated via perturbations are distributed very differently from instances in the COMPAS data. }
    % values generated by perturbing from a normal distribution
    %the synthetic data points generated from input perturbations are distributed significantly differently from the instances in the input data
    \label{fig:pca}
    \vspace{-0.2in}
\end{figure}

We consider a general problem setting in which explanations may be manipulated by adversaries.
We assume an \textit{adversarial model owner} wishes to train a biased model according to some notion of \textit{model bias} (e.g., demographic parity \cite{feldman2015Disparate}, recourse fairness \cite{karamirecoursesurvey2020}) and deploy this model in production.
However, a \textit{model auditor} will use explanations (e.g., LIME, counterfactual explanations) to audit the model for bias.
Thus, the adversarial model owner is incentivized to construct a model that is biased but explanations will indicate the model is unbiased, thereby decieving the auditor. 
% We will show that it is possible to train such models for both model agnostic feature attribution explanations and counterfactual explanations. 

\section{Manipulating Feature Attributions}

In this section, we describe how model agnostic feature attribution explanations are vulnerable to manipulation.

\subsection{Notion of Bias}

We assume that the adversary wishes to deploy a model that predicts the protected group as the negative outcome and non-protected group as the positive outcome.
The model auditor observes whether the feature attributions include the sensitive feature to see if bias occurs.
Thus, the adversary wants to devise a model that is biased in predictions but feature attributions do not highly rank sensitive features.

\subsection{Manipulation}

\paragraph{Intuition} As discussed in the previous section, LIME explains individual predictions of a given black box model by constructing local interpretable approximations (e.g., linear models) using perturbations in the vicinity of a given point. However, many instances generated by such perturbations are out of distribution because LIME \& SHAP rely on heuristically chosen perturbation functions.
%generate synthetic data points that are
% We carried out the following experiment. %data points resulting from input
% First, we perturb input instances using the approach employed by LIME (See previous section). 
We can see this running PCA on a combined dataset of instances and perturabations for LIME
% The two-dimensional scatter plot showing both the input as well as perturbed instances is in Figure~\ref{fig:pca}. %number of dimensions in the data
%
in Figure~\ref{fig:pca}. 
% the synthetic data points generated from input perturbations are distributed significantly differently from the instances in the input data. This result indicates that detecting whether a  data point is a result of a perturbation or not is not a challenging task, and thus approaches that rely heavily on these perturbations, such as LIME, can be \emph{gamed}. %perturbing input data typically generates out-of-distribution (OOD) samples that do not conform to the input data distribution. Furthermore, approaches such as LIME rely heavily on these perturbed instances (which are OOD samples) when constructing explanations. 

This intuition underlies our proposed approach.
By being able to differentiate between data points coming from the input distribution and instances generated via perturbation, an adversary can create an adversarial classifier (\emph{\scaffolding}) that behaves like the original classifier (perhaps be extremely discriminatory) on the input data points, but behaves arbitrarily differently (looks unbiased and \emph{fair}) on the perturbed instances, thus effectively fooling LIME or SHAP into generating innocuous explanations.
% Next, we formalize this intuition and explain our framework for building adversarial classifiers that can fool explanation techniques. 

% \paragraph{Building Adversarial Classifiers}
% Let $f$ be the biased classifier described earlier. The adversary would like to design a framework such that if and when end users generate explanations of this black box, post hoc techniques can be fooled into thinking that the model is innocuous.
% %
% %As discussed in the previous section, post hoc explanation techniques which rely on input perturbations such as LIME and SHAP have a major vulnerability that can be exploited by this adversary. 
% Recall that the real world data on which this classifier is likely to be applied follows a distribution $\mathcal{X}_{dist}$, and $\mathcal{X}$ is a set of $N$ data points sampled from this distribution that the adversary has access to. To fool the post hoc techniques, the adversary could design an adversarial classifier that exhibits biased behavior on instances sampled from $\mathcal{X}_{dist}$, and remain unbiased on instances that do not come from $\mathcal{X}_{dist}$. 
% Since the feature importances output by LIME and SHAP rely heavily on perturbed instances (which may typically be OOD samples, e.g. Figure~\ref{fig:pca}), the resulting explanations will make the classifier designed by the adversary look innocuous.

\paragraph{Training Objective} Assuming $\psi$ is a \emph{unbiased} classifier (e.g., makes predictions based on innocuous features that are uncorrelated with sensitive attributes), the adversarial classifier $e$ takes the following form: % (See Algorithm~\ref{alg:adv} in Appendix): 
\[
    e(x)= 
\begin{cases}
    f(x),& \text{if } x \in \mathcal{X}_{dist}\\
    \psi(x),              & \text{otherwise}
\end{cases}
\]
%$e(x) = f(x)$ whenever $x \in \mathcal{D}_{dist}$ and $e(x) = \psi(x)$ whenever $x \notin {D}_{dist}$ where $\psi(x)$.
% Algorithm~\ref{alg:adv} shows the complete pseudocode for the same. 
% In order to create such a classifier, we need to be able to decide whether a given data point $x$ comes from $\mathcal{X}_{dist}$ or not. %; we discuss how to build such a classifier next. 
where $\mathcal{X}_{dist}$ refers to the distribution of real world data. We train a separate classifier to determine if $x \in \mathcal{X}_{dist}$, simply by training a random forest to discriminate the perturbations generated from LIME and data from the training set.

% \paragraph{Detecting OOD Samples}
% To build a classifier $\texttt{is\_OOD}$ that detects if a given data point is an \emph{out-of-distribution} (OOD) sample ($\texttt{is\_OOD}(x) = \texttt{True}$) or not, we construct a new dataset from the instances in $\mathcal{X}$. 
% Specifically, we \emph{perturb} (more details in Experimental Evaluation) each instance $x$ in $\mathcal{X}$ to generate a new instance $x_p$. Let us denote the set of all the instances generated via perturbation as $\mathcal{X}_p$. The instances in $\mathcal{X}$ are then assigned the class label $\texttt{False}$ indicating that they are not OOD samples, while the instances in $\mathcal{X}_p$ are assigned the class label $\texttt{True}$ (indicating that they are OOD samples) unless they are already in $\mathcal{X}$. %unless they are very similar to the instances in $\mathcal{X}$. 
% We then train an off-the-shelf classifier on the combined dataset $\mathcal{X} \cup \mathcal{X}_p$ and their corresponding class labels (assigned as discussed above).  

\subsection{Experiments}

% In this section, we describe experimental setup and results on a successful manipulation using the COMPAS dataset~\cite{compas}.
\textbf{Setup} We run experiments on the COMPAS \cite{compas} dataset. 
% We split the data into training (90\%) and test (10\%) sets. 
% We build our framework using the training set and evaluate it on the test set. 
We construct $f$ to be a perfectly discriminatory classifier that uses a sensitive feature i.e., $f$ makes predictions purely based on race (if \emph{race = African American}, then predict 1, otherwise set predict 0).

% \textit{Perturbations \& OOD classifier} 
% For LIME, we generate perturbations for every data point in the data set $\mathcal{X}$ by adding random noise sampled from $\mathcal{N}(0,1)$ to each feature value. % of each data point. 
% For SHAP, we randomly choose a subset of features for each data point and mark their values as missing by replacing them with their corresponding values from background distribution\footnote{See \citet{lundberg17:a-unified} for more details.}.% about background distributions.}. 
% We choose the background distribution for SHAP to be cluster centers from kmeans with 10 clusters. We use these newly generated perturbations as well as the input data points to train the OOD classifier to detect OOD samples (See Proposed Framework section). We use the standard sklearn implementation of a random forest with 100 tree estimators as our OOD classifier.

% \textit{Unbiased Classifier $\psi$} 
We build the unbiased classifier $\psi$ by constructing synthetic \emph{uncorrelated features} that have zero correlation with sensitive attributes (e.g., race or gender). We experiment with one or two uncorrelated features. When we only have one uncorrelated feature in a particular experiment, $\psi$ solely uses that to make predictions (if uncorrelated feature = 1, then predict 1, else predict 0). On the other hand, when we have two uncorrelated features in an experiment, we  base the predictions on the xor of those two features. 
% Note that $\psi$ does not have to be restricted to always use synthetic uncorrelated features. It can also use any other existing feature in the data to make predictions.  
% We use default LIME tabular implementation without discretization, and the default Kernel SHAP implementation with kmeans with $10$ clusters as the background distribution.

\begin{figure}[tb]
    \centering
    % \begin{subfigure}{\columnwidth}
    %     \centering
        \includegraphics[width=1\columnwidth,clip,trim=20 0 25 0]{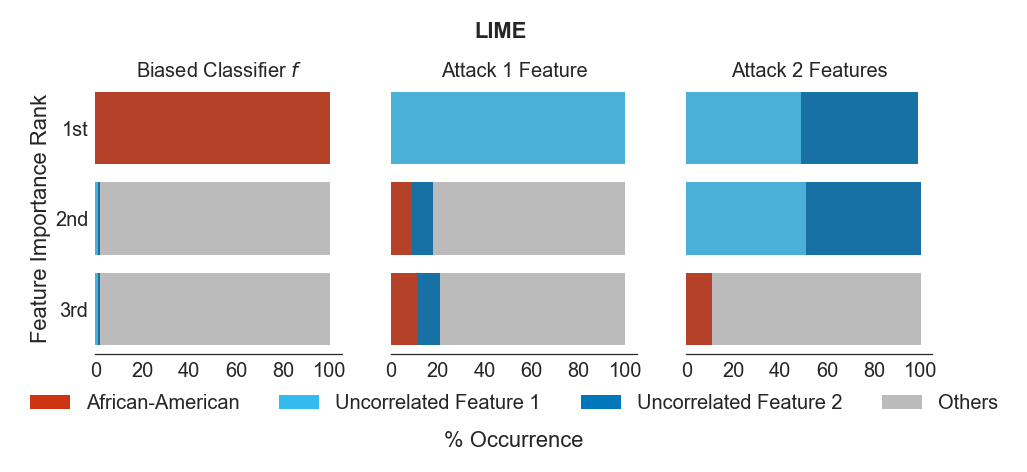}
        
        % \includegraphics[width=1\columnwidth,clip,trim=20 0 25 0]{figs/compas_shap_new.png}
    %     \caption{Ranking of features as per LIME explanations}
    %     \label{compas:lime}
    % \end{subfigure}
    % \begin{subfigure}{\columnwidth}
    %     \centering
    %     \includegraphics[width=1\columnwidth,clip,trim=20 0 25 0]{figs/compas_limeshap.png}
    %     \caption{Ranking of features as per SHAP explanations}
    %     \label{compas:shap}
    % \end{subfigure}
    %\caption{\textbf{COMPAS:} Proportion of each feature as top three most important ones according to LIME and SHAP, for the \emph{baseline} biased classifier ($f$) along with the ranking after 1-feature and 2-feature adversarial classifiers.
    \vspace{-0.2in}
    \caption{\textbf{COMPAS:} \% of data points for which each feature shows up in top 3 for the biased classifier $f$ (left), our adversarial classifiers (middle and right).
    % where $\psi$ uses one uncorrelated feature to make predictions (middle) and two (right). 
    % \sameer{shouldn't 2nd/3rd in left plots have grey/blue?} \dylan{Sophie adding currently}
    }
    \label{fig:compas_graph}
    \vspace{-0.2in}
    %adding vspace to remove weird spacing
\end{figure}

\paragraph{Results} To evaluate how successful our \emph{attacks} are on LIME, we compute the percentage of data points for which race and uncorrelated features show up in top 3 when features are ranked based on feature attributions output by LIME. 
% In figures~\ref{fig:compas_graph}, we show the results of these experiments.
From these results in figure~\ref{fig:compas_graph}, we can see that the attack is able to fool LIME and shift the feature importance to the uncorrelated feature $100\%$ of the time in both the one feature and two feature cases.  Last, the adversarial models make the same predictions as the biased model without the manipulation, indicating the model is still biased.
These results demonstrate the attack is effective.
% For \textbf{SHAP}, the attack is slightly less successful and the attack shifts feature importance from the senstive feature 71\% - 84\% of the time.
% when a single uncorrelated feature is used for the attack, the adversarial classifier $e$ successfully shifts the feature importance
%primary variable importance 
% from the sensitive feature in 84\% of data points  (Figure~\ref{fig:compas_graph} - bottom and middle). %, with the remaining 16\% of data points correctly identifying the bias feature. 
% When not correctly identified as the most important variable, the bias variable is frequently (98\% and 78\% for COMPAS and CC, respectively) identified as being of secondary importance. 
% When two uncorrelated features are used in the attack (for COMPAS and CC), the adversarial classifier is less successful in removing the bias feature from 1st place in the ranking, succeeding in only 71\% instances for CC respectively.
% This is due to SHAP's local accuracy property that ensures that feature attributions must add up to the difference between a given prediction and the average prediction for the background distribution. This property will tend to distribute feature attributions across several features when it is not possible to identify a single most informative feature. %Note also that SHAP percentage rankings do not always sum to 1, as it often assigns features a zero importance. %Detailed results can be found in Table \ref{table:COMPAS} and Table \ref{table:CC}.

\section{Manipulating Counterfactual Explanations}

In this section, we show how counterfactual explanations that hill-climb are vulnerable to manipulation.

\subsection{Notion of Bias}

As the notion of \textit{model bias}, we assume that the adversarial model owner want to deploy a model that is \textit{recourse biased} \cite{karamirecoursesurvey2020, guptaequalrecourse}.  Meaning, the recourse costs (as determined by the distance function $d$) is much lower for the non-protected group than the protected group, i.e.,     $\mathbb{E}_{x \sim \dataset_\pro^\negout} \left[ \cost \left( \datainstance, \alg(\datainstance) \right) \right] 
    \gg \mathbb{E}_{x \sim \dataset_\unpro^\negout} \left[ \cost \left( \datainstance, \alg(\datainstance) \right) \right]$.  Thus, the adversary is incentivized to create a model that exhibits equal recourse costs between groups on the data distribution but can produce much lower recourse for the non-protected group.

\subsection{Manipulation}

\begin{figure}
    \centering
    \includegraphics[width=.8\columnwidth]{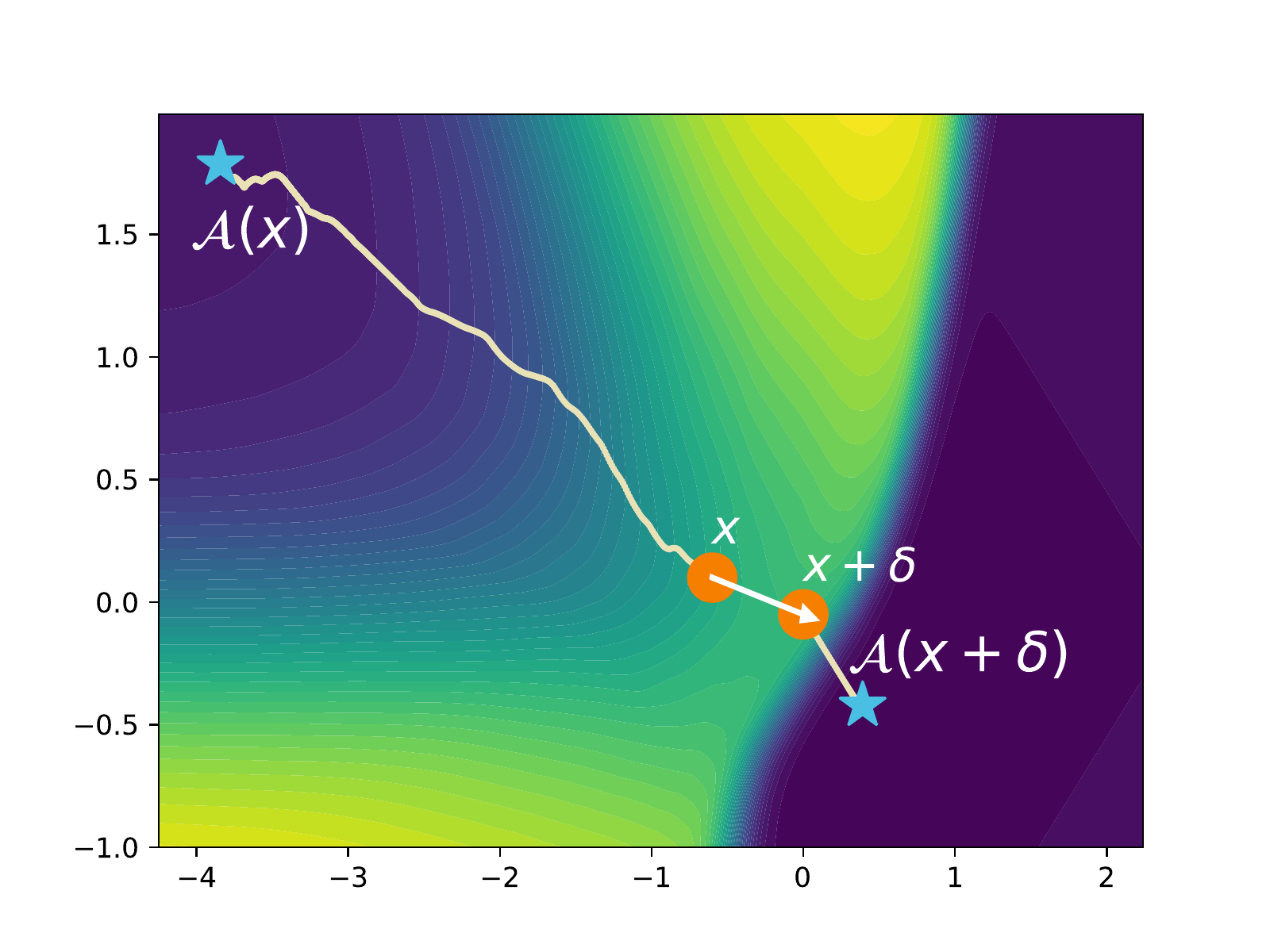}
    \caption{\textbf{Adversarial Model loss surface:} the recourse found for $\datainstance$ is \textit{higher cost} than $\datainstance+\key$ because the local minimum initialized at $\datainstance$ is \textit{farther} than the minimum starting at $\datainstance + \key$, demonstrating the problematic behavior of counterfactual explanations.}
    \label{fig:intuition}
\end{figure}

\paragraph{Intuition} Because counterfactual explanations that hill-climb use gradient descent, it is possible to design a model where the counterfactual explanations converge to drastically different counterfactuals when a small perturbation $\key$ (a vector of the same dimensions as $\datainstance$) is added to instances.
We introduce an adversarial objective that leverages this insight.
This objective encourages the model to return equal cost recourse between groups, indicating fairness in the counterfactuals to the auditor.
However, counterfactual explanations return much lower cost counterfactuals when the perturbation $\key$ is added.
In this way, the adversary can simply add the perturbation $\key$ to generate low cost recourse.

We show such an adversarial model in Fig~\ref{fig:intuition} trained on a toy data set. The counterfactual found for the perturbed instance $\datainstance + \key$ is \textit{closer} to the original instance $\datainstance$.  
This result indicates that the counterfactual found for the perturbed instance $\datainstance + \key$ is \textit{easier to achieve} than the counterfactual for $\datainstance$ found by Wachter's algorithm!
Intuitively, counterfactual explanations that hill-climb the gradient are susceptible to this issue because optimizing for the counterfactual at $\datainstance$ versus $\datainstance + \key$ can converge to different local minima. 

\paragraph{Training Objective for Adversarial Model}
\label{sec:manip_cf_exps}
% In this situation, a bad actor designs a model that returns lower distance counterfactuals for the not protected group than the protected group. Also, the bad actor wants to pass external audits on the counterfactual disparity between groups. Thus, the actor manipulates the model to return no counterfactual disparity when audited.  
% To produce such an adversarial model, the actor trains a model with no counterfactual disparity.  However, the model produces lower distance counterfactuals when the actor adds an imperceptible perturbation for certain groups.  Consider the loan provider. The counterfactual algorithm indicates that men and women who are denied loans must earn on average $\$100$ more to be approved. When ever the bad actor wants to generate a low recourse counterfactual for the men, they can add the perturbation.  This attack is realistically viable.  The model would appear counterfactual fair to any auditor who examines the counterfactuals returned by the model.  The attacker could add small perturbations to the not protected group to generate low distance counterfactuals in production and use the same model all the while.
%
We define our objective using the combination of the following terms:
\begin{itemize}[nosep,noitemsep,leftmargin=12pt]
    \item \emph{Fairness:} The counterfactual algorithm $\alg$ is unbiased for model $\model$.
    \item \emph{Unfairness:} A perturbation vector $ \key \in \mathbb{R}^d$ leads to lower cost recourse when added to  non-protected data, leading to unfairness.
    % i.e., $\mathbb{E}_{x \sim \dataset_\pro^\negout} \left[ \cost \left( \datainstance, \alg(\datainstance) \right) \right] \gg  \mathbb{E}_{x \sim \dataset_\unpro^\negout} \left[ \cost \left( \datainstance, \alg(\datainstance + \key) \right) \right]$.
    \item \emph{Small perturbation:} Perturbation $\delta$ should be \emph{small}. i.e. we need to minimize $\mathbb{E}_{\datainstance \sim \dataset_\unpro^\negout}\cost (\datainstance, \datainstance + \key )$.
    \item \emph{Accuracy:} We should minimize the classification loss $\mathcal{L}$ (such as cross entropy) of the model $\model$ .
    % \item \emph{Counterfactual:} $(\datainstance+\key)$ should be a counterfactual, i.e. minimize $\mathbb{E}_{\datainstance \sim \dataset_\unpro^\negout} \; (f(\datainstance + \key) - 1)^2$.
\end{itemize}
This combined training objective is defined over both the parameters of the model $\mparams$ and the perturbation vector $\key$.
This objective is complicated by the fact that it involves $\mathcal{A}$, a black-box counterfactual explanation approach.
 Because optimizing the objective involves solving a bi-level optimization problem, we can compute the gradients for $\alg$ through implicit differentiation \cite{bilevel_gould}. 

\subsection{Results}

We consider four different counterfactual explanation algorithms as the choices for $\alg$ that hill-climb the counterfactual objective.
We use \textbf{\em Wachter}'s Algorithm~\cite{wachter}, Wachter's with elastic net sparsity regularization (\textbf{\em Sparse Wachter}; variant of \citet{NEURIPS2018_c5ff2543}), \textbf{\em DiCE} \cite{mothilal2020dice}, and Counterfactual's Guided by \textbf{\em Prototypes}~\cite{proto20}. %\sameer{same cite for two of these?} \dylan{probably should include a bit more description about these for readers... differences etc...} 
% These counterfactual explanations are widely used to compute recourse and assess the fairness of models \cite{karamirecoursesurvey2020, reviewcounterfactuals, surveycfs}.
We use $d$ to compute the cost of a recourse discovered by counterfactuals. %, %, must decide what notions of distance we will use. %  the effectiveness of the counterfactuals, and consequently our attacks
% we use 
% the $\ell_1$ distance 
%for the $\ell_p$ distance\sameer{there's no $\ell_p$ anywhere earlier, right?} 
%in the algorithms %because it is representative of what someone would actually have to do to achieve the counterfactual. 
% in t
% the distance function $d$.
% , which also includes algorithm-specific additional counterfactual desiderata. % in each algorithm during assessment. %\sameer{basically what is used in objectives? unclear}  as the sum of all the  
% We use the official DiCE implementation\footnote{\url{https://github.com/interpretml/DiCE}}, and reimplement the others. 
We use the \textbf{Communities \& Crime} data set to train the adversarial models.
We use a feedforward neural network as the adversarial model. 
% consisting of $4$ layers of $200$ nodes with the ReLU activation function, the Adam optimizer, and using cross-entropy as the loss $\mathcal{L}$. 

We evaluate the effectiveness of the manipulated models across counterfactual explanations using both the disparity of the average recourse cost for protected and non-protected groups (i.e., the model seems fair to auditors) and
% We also measure the average costs (using $d$) for the non-protected group and the non-protected group perturbed by $\key$.
% (1) the average cost (using $d$) between each data point and its counterfactual, computed on protected, not-protected group,.
the ratio between the non-protected group and the non-protected group perturbed by $\key$ (i.e., model can generate low cost recourse for the non-protected group), which we denote as the \textit{cost reduction}.
% these costs as metric for success of manipulation, % the \textit{cost ratio} due to $\key$,
% % \dylan{cost gap could be a better name? recourse gap? Cost ratio} 
% \begin{equation}
%     \textrm{Cost reduction} := \frac{ \mathbb{E}_{x\sim \dataset_\unpro^\negout} \left[ d(\datainstance, \alg(\datainstance) ) \right] }{\mathbb{E}_{x\sim \dataset_\unpro^\negout} \left[ d(\datainstance, \alg(\datainstance + \key)) \right] }.
%     \label{eq: reduction}
% \end{equation}
If the manipulation is successful, we expect the cost reduction to be high. 
% We measure the difference between mean counterfactual distance between the protected and not-protected groups to assess counterfactual fairness (i.e., equation \ref{eq: counterfactual fairness}). 
% (3) We evaluate $||\key||_1$ and model accuracy to make sure the perturbation is small and the model is accurate.

We provide the results for both datasets in Table~\ref{tab:effectiveness}.
All the models trained have within 1\% accuracy of baseline neural networks trained on the data without the manipulation (not in table), indicating they are still accurate. 
The disparity in counterfactual cost on the unperturbed data is very small in most cases, indicating the models would appear counterfactual fair to the auditors.
% compared to the baseline 
At the same time, we observe that the cost reduction in the counterfactual distances for the non-protected groups after applying the perturbation $\key$ is quite high, indicating that lower cost recourses are easy to compute for non-protected groups.
These results show the adversarial models are successful.

\begin{table}[tb]
\small
\centering
\caption{\textbf{Recourse Costs of Manipulated Models}: Counterfactual algorithms find similar cost recourses for both subgroups, however, give much lower cost recourse if $\key$ is added.} %protected and non-protected 
\label{tab:effectiveness}
\begin{tabular}{l rrrr} 
\toprule 
& \multicolumn{4}{c}{\bf Communities and Crime}\\
 \cmidrule(lr){2-5}
%  \cmidrule(lr){6-9}
&  \multicolumn{1}{c}{\bf Wach.} & \multicolumn{1}{c}{\bf S-Wach.} & \multicolumn{1}{c}{\bf Proto.} &  \multicolumn{1}{c}{\bf DiCE}  \\ 
\midrule %\vspace{1mm}
% Model & Baseline & Adv. & Baseline & Adv. & Baseline & Adv. & Baseline & Adv. \\ 
% \midrule
Protected &  35.68 &  54.16  &  22.35 &  49.62 \\  
Non-Protected  & 35.31 & 52.05 & 22.65 & 42.63 \\ 
\em Disparity & \em 0.37 & \em 2.12&  \em 0.30 &  \em 6.99  \\
\addlinespace
Non-Protected$+\key$ & 1.76 & 22.59 & 8.50 & 9.57  \\ 
\em Cost reduction & {\em 20.1$\times$} & {\em 2.3$\times$} & {\em 2.6$\times$} & {\em 4.5$\times$} \\
\bottomrule
\end{tabular}
\vskip -2mm
\end{table}%

% \section{Related Work}

\section{Conclusion}

In this work, we demonstrate how two broad classes of post hoc explanations are vulnerable to manipulation. These results raise serious concerns around the usefulness of explanations as a tool to endow trust in machine learning models. Particularly in high stakes settings, these methods can easily be gamed by bad-actors to hide undesirable aspects of models. Further, these results highlight the need to develop machine learning methods that are robust to manipulation in the future, if we are to trust explanation methods.

\section{Acknowledgments}

We would like to thank the anonymous reviewers for their insightful feedback. This work was supported in part by the NSF awards \#IIS-2008461 and \#IIS-2008956,  Google, Amazon, and the Hasso-Plattner Institut. The views expressed are those of the authors and do not reflect the official policy or position of the funding agencies.

\bibliography{bib}
\bibliographystyle{icml2021}

\end{document}